\documentclass[conference]{IEEEtran}
\IEEEoverridecommandlockouts
\usepackage{cite}
\usepackage{amsmath,amssymb,amsfonts}
\usepackage{algorithmic}
\usepackage{graphicx}
\usepackage{textcomp}
\usepackage{xcolor}
\usepackage{algorithm} 
\usepackage{fancyhdr}
\usepackage{url}

\usepackage[a4paper, total={184mm,239mm}]{geometry}
\def\BibTeX{{\rm B\kern-.05em{\sc i\kern-.025em b}\kern-.08em
    T\kern-.1667em\lower.7ex\hbox{E}\kern-.125emX}}
\begin{document}

\title{VW-SDK: Efficient Convolutional Weight Mapping Using Variable Windows for Processing-In-Memory Architectures\\

}

\author{\IEEEauthorblockN{Johnny Rhe\IEEEauthorrefmark{1}, Sungmin Moon\IEEEauthorrefmark{1}, and Jong Hwan Ko\IEEEauthorrefmark{2}}
\IEEEauthorblockA{\IEEEauthorrefmark{1}Department of Electrical and Computer Engineering, Sungkyunkwan University, Suwon, South Korea \\
\IEEEauthorblockA{\IEEEauthorrefmark{2}College of Information and Communication Engineering, Sungkyunkwan University, Suwon, South Korea\\
\{djwhsdj, sang6989, jhko\}@skku.edu}  }
}

\maketitle
\thispagestyle{fancy}
\lhead{Accepted as a conference paper at Design, Automation \& Test in Europe Conference \& Exhibition (DATE) 2022}
\rhead{}
\cfoot{}

\begin{abstract}
With their high energy efficiency, processing-in-memory (PIM) arrays are increasingly used for convolutional neural network (CNN) inference. In PIM-based CNN inference, the computational latency and energy are dependent on how the CNN weights are mapped to the PIM array.
A recent study proposed shifted and duplicated kernel (SDK) mapping that reuses the input feature maps with a unit of a parallel window, which is convolved with duplicated kernels to obtain multiple output elements in parallel.
However, the existing SDK-based mapping algorithm does not always result in the minimum computing cycles because it only maps a square-shaped parallel window with the entire channels.
In this paper, we introduce a novel mapping algorithm called variable-window SDK (VW-SDK), which adaptively determines the shape of the parallel window that leads to the minimum computing cycles for a given convolutional layer and PIM array.
By allowing rectangular-shaped windows with partial channels, VW-SDK utilizes the PIM array more efficiently, thereby further reduces the number of computing cycles.
The simulation with a 512$\times$512 PIM array and Resnet-18 shows that VW-SDK improves the inference speed by 1.69$\times$ compared to the existing SDK-based algorithm.
\end{abstract}


\section{Introduction}
Convolutional neural networks (CNNs) have advanced the state-of-the-art machine learning algorithm across various computer vision applications with increased layers and weight parameters.
The inference of large CNN models on the traditional von Neumann architecture inevitably leads to a significantly large energy consumption due to extensive data movement between the processor and memory for the weight parameter access.
Recent studies eliminated the need for weight movement during inference by using processing-in-memory (PIM) arrays that can perform computation in the memory cell storing the weight element. 

For PIM-based CNN inference, the kernel and input elements of convolutional layers need to be mapped to the memory cells and input ports of the PIM array. However, the size of recent PIM arrays is not sufficient to map the entire layer of CNN models \cite{pipelayer}, requiring multiple mapping and computing cycles to obtain the complete output feature maps of a certain layer.
More computing cycles will lead to higher energy consumption because of an increase in the latency and the analog-digital/digital-analog conversions \cite{efficient, ADC}.
Therefore, for fast and efficient PIM-based CNN inference, it is crucial to optimally map the convolutional layers into a PIM array.

\begin{figure}[t]
\centering
\includegraphics[trim={0cm 0cm 0cm 0cm}, clip, width=0.9\linewidth]{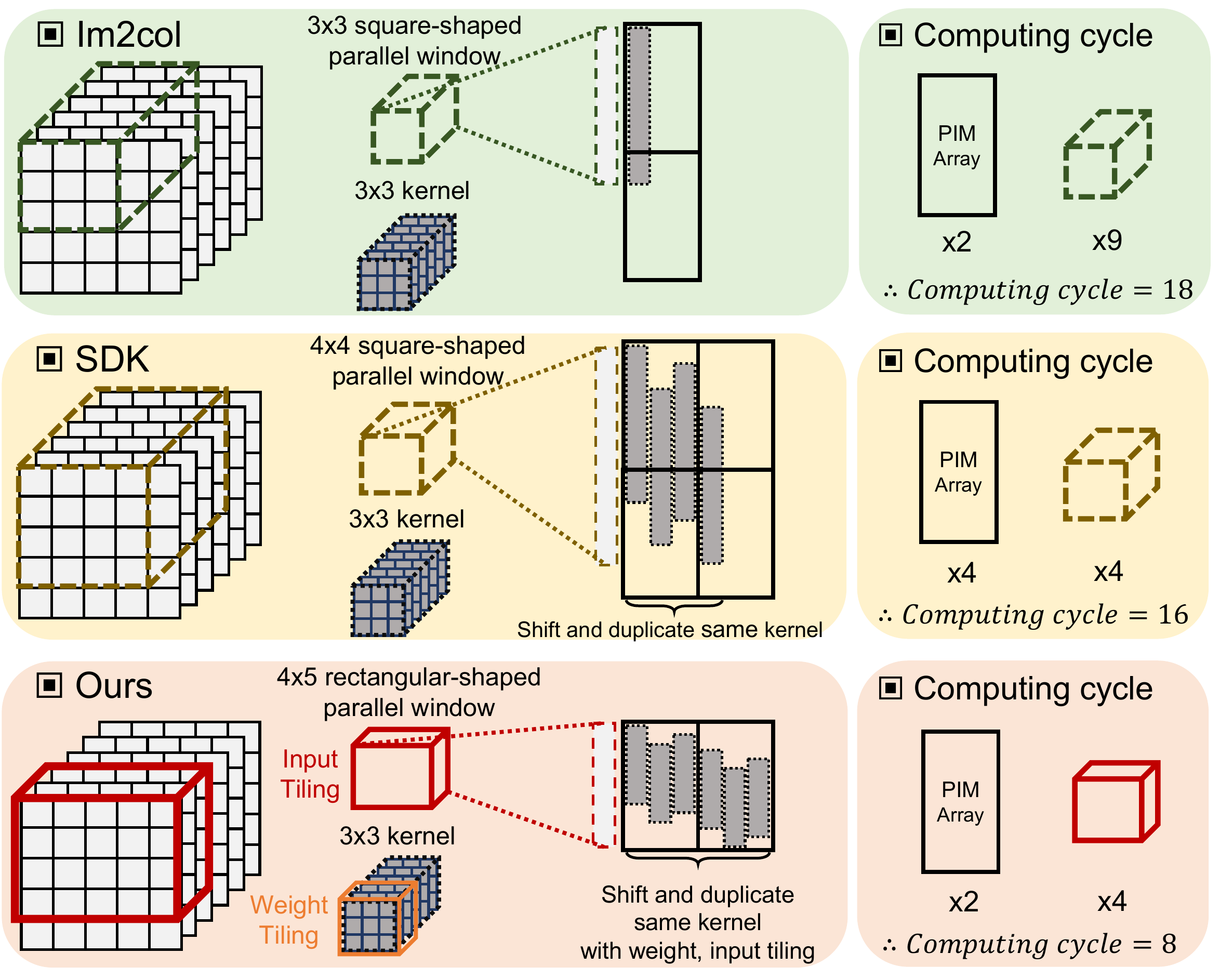}
\caption{Conventional methods and the proposed one for mapping of convolutional weights on a PIM array.}
\label{fig:fig1}
\end{figure}

\begin{figure*}[t]
\centering
\includegraphics[trim={0cm 0.5cm 0cm 0cm}, clip, width=0.85\linewidth]{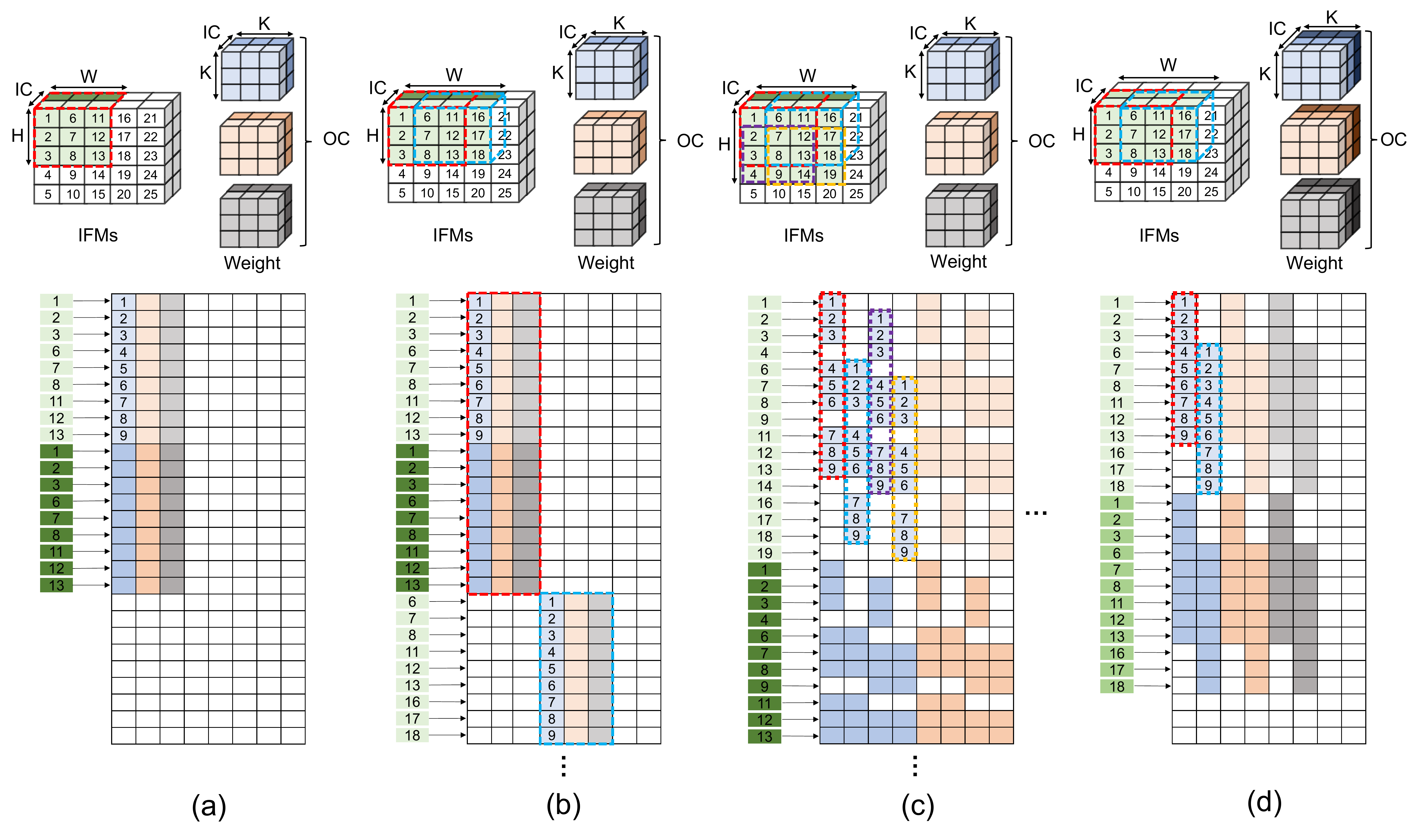} 
\caption{Methods for mapping the convolutional weights into a PIM array. (a) Im2col. (b) Sub-matrix duplication. 
(c) SDK. 
(d) The proposed VW-SDK.
}
\label{fig:mapping}
\end{figure*}

Image to column (im2col) \cite{im2col}, one of the early mapping methods, decomposes the kernel weights into the column of a PIM array and performs vector-matrix multiplications with the input data (Fig. \ref{fig:fig1} (top)).
However, im2col cannot fully utilize a given PIM array because this mapping does not reuse elements of input data when the kernels are sliding over the input feature map \cite{mixed}.
To address this problem, a previous study \cite{optimizing} proposed duplicating the kernel weights and placing them inside the sub-array to obtain the multiple outputs simultaneously.
In addition, another study proposed shift and duplicate kernel (SDK) mapping \cite{efficient} that reuses the input data with a unit of a parallel window that is shared across the duplicated kernels, to enhance the array utilization (Fig. \ref{fig:fig1} (middle)).
However, the SDK mapping algorithm utilizes only the square-shaped parallel window, significantly restricting the chance for further optimization. 
Also, as it duplicates the entire channel of the kernels, this mapping scheme cannot reduce the total computing cycles when the PIM array is not enough to unroll the duplicated and shifted kernels.
In this paper, we propose a novel method called variable-window SDK (VW-SDK) for efficient mapping of convolutional layers into a PIM array.
By forming variable parallel windows with partial channels and rectangular-shapes, the proposed method further enhances the utilization of a PIM array (Fig. \ref{fig:fig1} (bottom)).
Based on VW-SDK, we also propose an algorithm to find the optimal shape of a parallel window that results in the minimum computing cycles for a given PIM array.
The simulation results show that VW-SDK with a 512$\times$512 PIM array improves the computing speed by 4.67$\times$ and 1.69$\times$ for Resnet-18, compared to the im2col and SDK-based algorithms, respectively.

\section{BACKGROUND}

\subsection{Mapping of Convolutional Layers in PIM Architectures}

In order to perform convolution operations on a PIM array, elements in the kernels and input feature maps need to be respectively mapped to the cells and input ports of the array \cite{efficient}. 
In general, the weight elements of each 3D kernel are unrolled to the memory cells of the same column. 
Then, the input voltage as a part of input feature maps (IFMs) is assigned to each row to generate the accumulated current vector, which is a part of the output feature maps (OFMs).

As general PIM array sizes are not enough to assign the entire inputs and weights of recent complex CNN layers \cite{pipelayer}, multiple mapping and computing cycles are needed to obtain the complete OFMs of a given CNN.
More computing cycles increase the inference latency, thereby resulting in the inference energy increase.
More computing cycles also increase the number of data conversions required every cycle between the analog signal and digital input/output data \cite{efficient}, which are known to cost more than 98\% of the total PIM energy consumption \cite{ADC}.
Therefore, efficient mapping of the convolutional layers into a PIM array is crucial for fast and efficient PIM-based CNN inference.
A few studies have proposed schemes that map convolutional layers to a PIM array as follows.

{\bf Im2col.}{\bf}
Image to column (Im2col) \cite{im2col} is the simplest mapping method that unrolls each kernel with size K$\times$K$\times$IC into a column, where IC is the number of input channels and K is the size of the kernel.
In this scheme, a kernel-sized window in an IFM is convolved with the kernel to generate the output elements, as shown in Fig. \ref{fig:mapping}(a).

{\bf Sub-Matrix Duplication.}{\bf}
The sub-matrix duplication \cite{optimizing} places the duplication of kernels, allowing more IFMs to be operated at the same time, as shown in Fig. \ref{fig:mapping}(b).
Because this mapping enables computation of multiple output elements in one cycle, it can reduce the computing cycles than Im2col when the array space is sufficient to unroll the duplicated kernels.

{\bf SDK.}{\bf}
Shift and duplicate kernel (SDK) mapping \cite{efficient} reuses the input data and weights by shifting and duplicating kernels on adjacent columns, as shown in Fig. \ref{fig:mapping}(c).
This method forms the parallel window, a set of windows sharing the part of kernels, to obtain multiple OFM elements at one cycle.
By increasing the utilization of the PIM array, the computing cycles can be reduced compared to Im2col or sub-matrix duplication mapping. 
The SDK mapping algorithm proposed in \cite{efficient} duplicates the entire channel of the kernels in the unit of square number, to form the square-shaped parallel window.
When the duplicated kernels are significantly large to be mapped to a given PIM array, the number of computing cycles cannot be reduced compared to the im2col mapping.

\subsection{Calculation of the Number of Computing Cycles}
\label{Calculation Of Cycles}
\begin{figure}[!h]
\centering
\includegraphics[trim={0cm 0.5cm 0cm 0.3cm}, clip, width=0.8\linewidth]{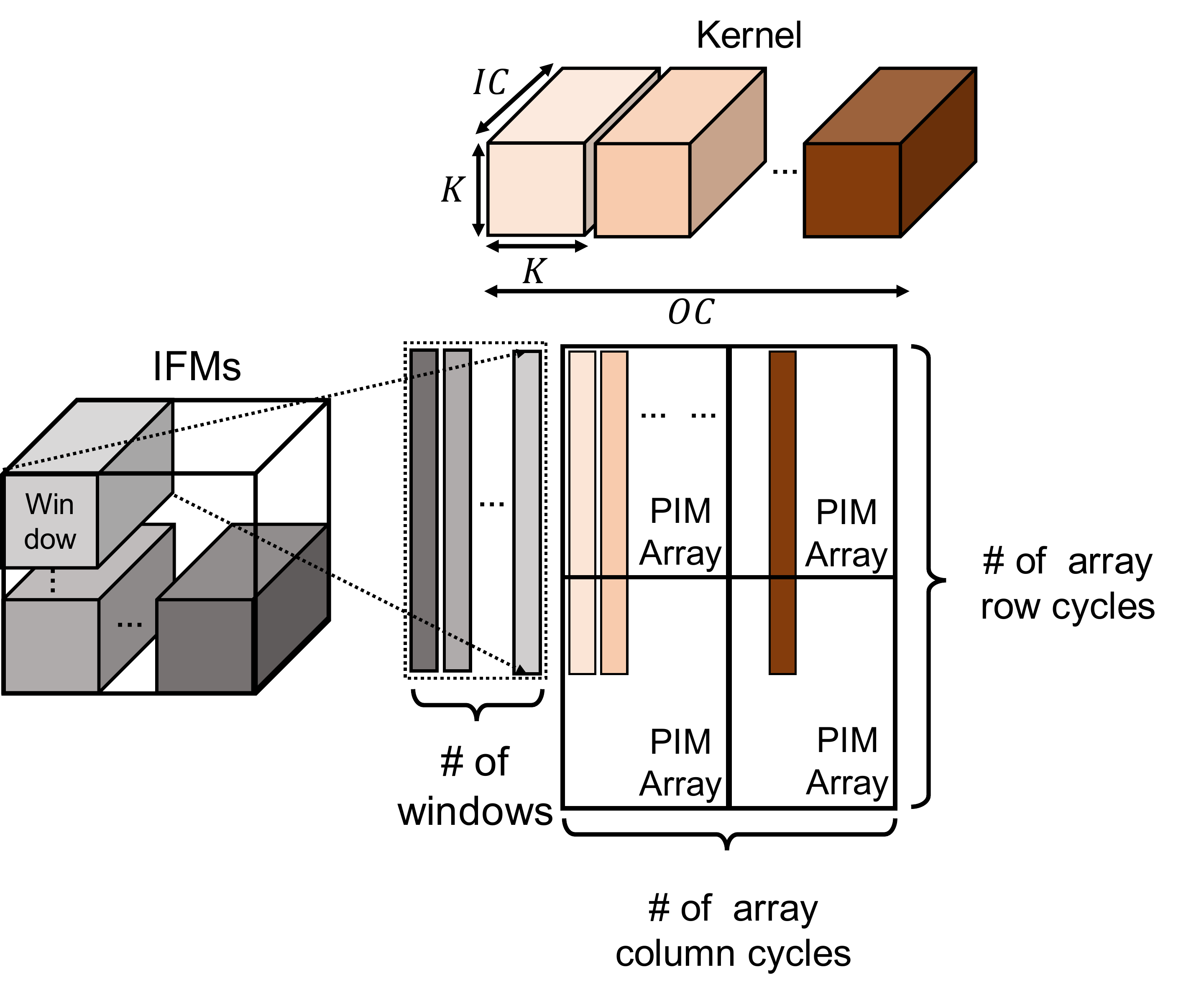} 
\caption{Illustration of explaining how the number of computing cycles is calculated.}
\label{fig:howtoCC}
\end{figure}

In a convolutional layer, a window slides over the entire IFMs and convolves with the kernel to extract OFMs. 
If a given PIM array is large enough to unroll the window-sized input and entire kernels, the number of computing cycles required to compute the given convolutional layer is equal to the number of windows in the IFMs.
Otherwise, additional computing cycles are needed to obtain the entire OFMs, as shown in Fig. \ref{fig:howtoCC}.
The additional cycles can be divided into two components: the cycles that require more rows (the array row cycles or AR cycles) and the cycles that require more columns (the array column cycles or AC cycles).
The AR cycles are defined as the cycles required to compute all the multiplications between the elements in one kernel and the corresponding window.
Similarly, the AC cycles are defined as the cycles required to generate partial products of the entire output channels.
By using these terms, the total number of computing cycles can be expressed as follows:
\begin{equation}
\scriptstyle
    computing\ cycles\ =\ N\ of\ windows\ \times\ AR\ cycles\ \times\ AC\ cycles,
\label{Eq:computingcycle}
\end{equation}
where AR cycles and AC cycles are $\lceil{\frac{PW \times PW \times IC}{2^{X}}}\rceil$ and $\lceil{\frac{OC \times NW_{P} }{2^{Y}}}\rceil$, respectively; $\lceil\cdot\rceil$ is the ceil function; PW is the size of the parallel window; OC is the number of output channels; $2^X$ and $2^Y$ are the number of rows and columns of the PIM array, respectively.
In this equation, when the number of windows in the parallel window (referred as $NW_{P}$) is 1 (PW$\times$PW is equal to K$\times$K), the SDK mapping can be considered as the im2col mapping.

\section{MOTIVATION}
\label{Motivation}

\subsection{Limited Array Size}
\label{motivation_limited array size}

\begin{figure}[!h]
\centering
\includegraphics[trim={0.1cm 0.1cm 0.1cm 0.1cm}, clip, width=0.85\linewidth]{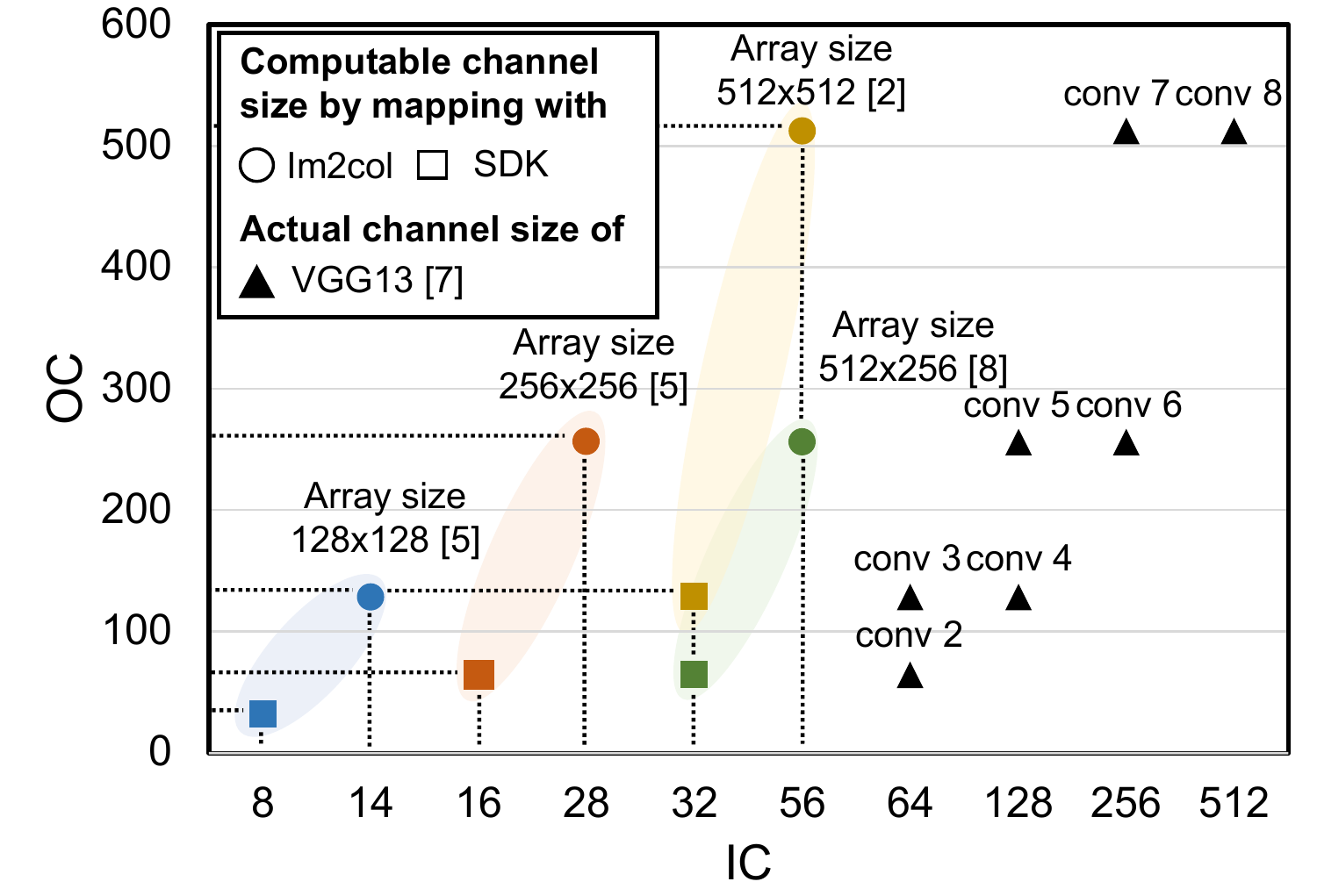} 
\caption{The number of input/output channels that can be mapped at one cycle by each mapping method for various PIM array sizes (the circle represents im2col; the square represents SDK with 4$\times$4 parallel windows). The triangles are the input/output channels of VGG-13 layers used in \cite{vggnet}. The figure indicates that the conventional mapping methods cannot map the entire channels of general conv layers into contemporary PIM arrays at one cycle.
}
\label{fig:limitedarray}
\end{figure} 
In contemporary CNNs, convolutional layers generally include many channels ranging from 64 to 512 \cite{vggnet}.
However, current PIM array sizes are significantly limited (e.g. 128$\times$128 \cite{mixed}, 256$\times$256 \cite{mixed}, 512$\times$512 \cite{efficient}, 512$\times$256 \cite{512x256}) to map the entire channels of those layers, as shown in Fig. \ref{fig:limitedarray}.
Nevertheless, the existing mapping algorithms are designed to map the entire channels to an array, thereby significantly restricting the computable channel size. 
The computable channel size is further restricted when SDK is used because more rows are needed to unroll the parallel window and more columns are needed to duplicate the kernels.
Therefore, the mapping algorithm must consider dividing channels into several tiles to map large layers on a size-limited PIM array using an efficient SDK method.

\subsection{Optimal Shape of Parallel Window}
\label{section:optimal}
\begin{figure}[!h]
\centering
\includegraphics[trim={0cm 0.7cm 0cm 0cm}, clip, width=0.95 \linewidth]{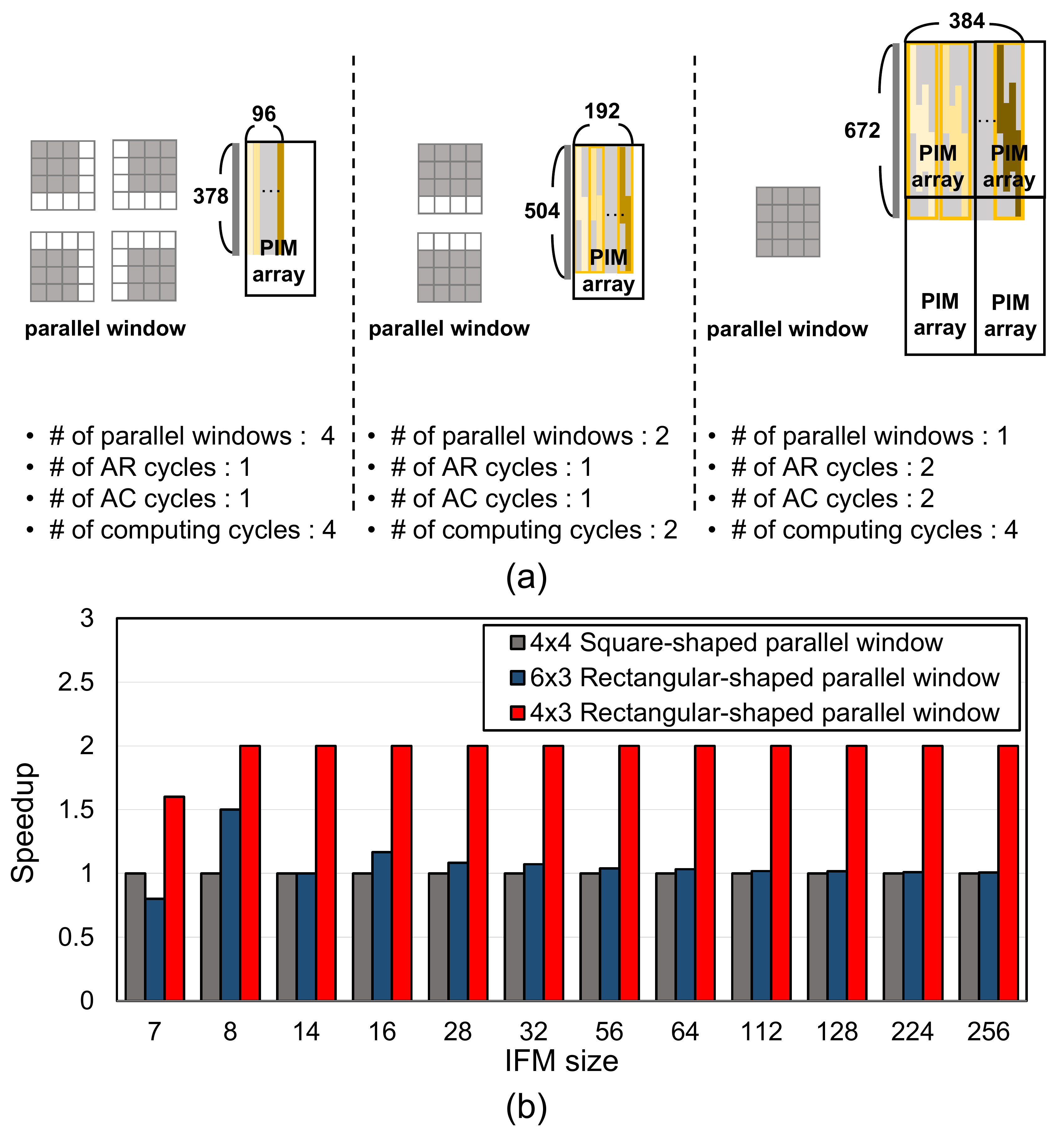}
\caption{
(a) shows how the number of computing cycles of im2col, 4$\times$3 parallel window and 4$\times$4 parallel window is calculated. This example assumes the sizes of the PIM array, kernel, IC, OC are 512$\times$256, 3$\times$3, 42, and 96, respectively.
(b) compares the speedup of square-shaped parallel window with rectangular-shaped parallel window when the size of IFMs increases. The x-axis is based on the image size used by VGGNet \cite{vggnet} (e.g., 14 indicates a parallel window size of 14$\times$14). 
}
\label{fig:optimal}
\end{figure}

The existing SDK-based algorithm searches for the parallel windows with a square shape to minimize the required computing cycles. 
However, it is not always optimal because the number of computing cycles may not be reduced due to the increase of AR and AC cycles.
Although SDK with the square-shaped parallel window can reduce the number of parallel windows compared to im2col, it is not always optimal because the number of computing cycles may not be reduced due to the increase of AR and AC cycles, as shown in Fig. \ref{fig:optimal}(a). 
In contrast, if we extend the search space into rectangular shapes, we can find the optimal window shape that leverages the given PIM array without increasing AR and AC cycles, resulting in the minimum computing cycles. 
Fig. \ref{fig:optimal}(b) shows that the rectangular-shaped parallel window leads to less computing cycles than the square-shaped parallel window under various IFM sizes. 
In particular, a 4$\times$3 rectangular-shaped parallel window achieves $\sim$ $2\times$ speedup compared to the 4$\times$4 square-shaped parallel window.
Therefore, the mapping algorithm should consider the rectangular shape when searching for the optimal shape of the parallel window.

\section{Proposed Method}
\subsection{SDK with Variable Parallel Window}
\begin{figure}[!h]
\centering
\includegraphics[trim={0cm 0.5cm 0cm 0.5cm}, clip, width=0.9\linewidth]{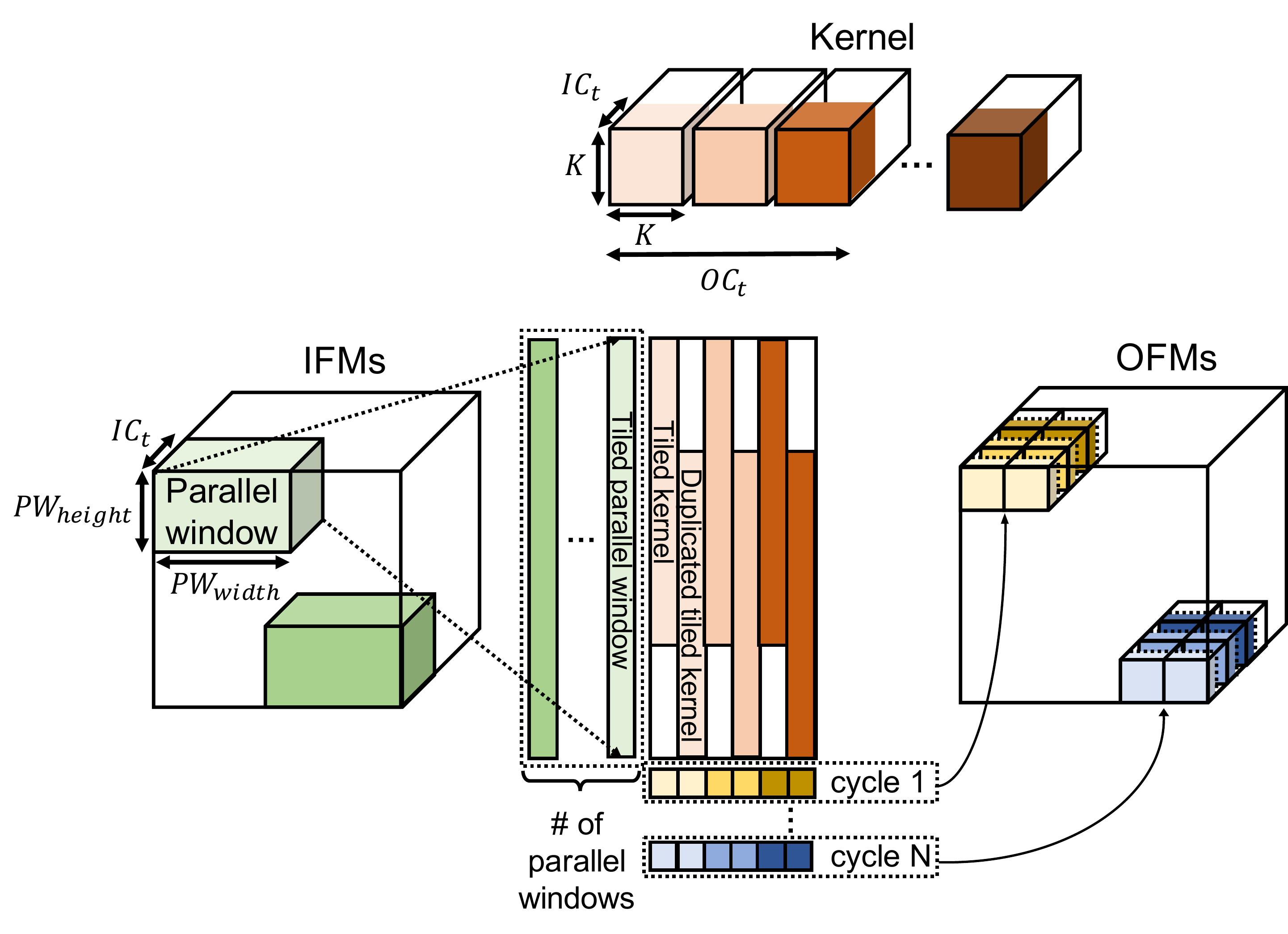}
\caption{Concept of the VW-SDK mapping method.}
\label{fig:VW-SDK}
\end{figure} 

Based on the SDK method, we propose a variable-window SDK (VW-SDK) method that allows the parallel windows to be variable in terms of both the channel size and the window shape, as shown in Fig. \ref{fig:mapping}(d) and Fig. \ref{fig:VW-SDK}.
First, instead of mapping the entire channels of a convolutional layer, the proposed method maps a part of the channels in a cycle.
As the existing SDK mapping method maps the entire channels, it cannot utilize large parallel windows that leads to higher utilization, especially when the array size is limited. 
The proposed method reduces the channel size to allow mapping of the kernel with the optimal parallel window, resulting in minimum computing cycles.
Although the proposed method computes only a part of channels in the OFMs at a single cycle, the total computing cycles can be reduced as more output elements per channel can be simultaneously obtained.

Second, the proposed method utilizes rectangular-shaped parallel windows.
By allowing various shapes of the parallel window, the computing cycles can be further minimized, as mentioned in Section \ref{section:optimal}.
With these concepts, VW-SDK leverages the given PIM array by forming the optimal parallel window for each convolutional layer.

\subsection{Optimal Parallel Window Search}

When calculating the number of computing cycles with SDK mapping, the number of windows can be substituted by the number of parallel windows in \eqref{Eq:computingcycle} as follows:

\begin{equation}
\scriptstyle
    computing\ cycles\ =\ N\ of\ PWs\ \times \ AR\ cycles\ \times \ AC\ cycles.
\label{Eq:computingcycle2}
\end{equation}

The number of parallel windows depends on the size of the parallel window and IFMs as follows: 

\begin{figure}[t]
\centering
\includegraphics[trim={0.2cm 0.4cm 0cm 0.1cm}, clip, width=0.95 \linewidth]{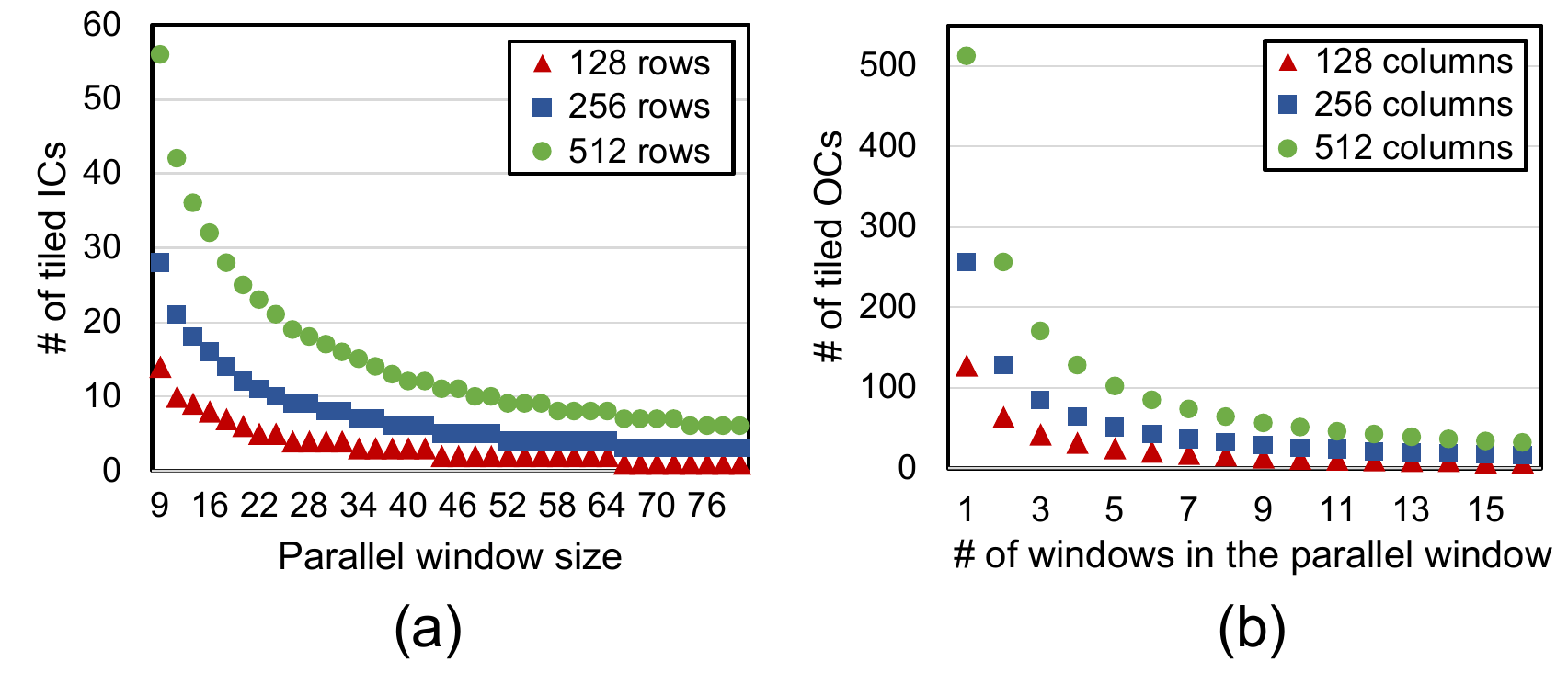}
\caption{
With a different PIM array size, 
(a) shows the number of tiled ICs when the area of the parallel window increases.
(b) shows the the number of tiled OCs when the number of windows in the parallel window increases.}
\label{fig:IC1C}
\end{figure}

\begin{equation}
\scriptstyle 
    N\ of\ PWs\ =\ 
    \left(\big\lceil\frac{I_{w}-PW_{w}}{PW_{w}-K_{w}+1}\big\rceil+1\right) \
    \times \ \left(\big\lceil\frac{I_{h}-PW_{h}}{PW_{h}-K_{h}+1}\big\rceil+1\right),
\label{Eq:inputrow}
\end{equation}
where I is the size of the IFM. 
When the size of the parallel window increases, we must decrease the number of channels to be mapped to a given PIM array because the number of rows of the PIM array is limited, as shown in Fig. \ref{fig:IC1C}(a). Thus, the maximum size of the tiled ICs for a certain parallel window can be obtained by
\begin{equation}
\scriptstyle
   IC_{t}\ =\ \big\lfloor\frac{2^X}{PW_{w}\times PW_{h}}\big\rfloor,
\label{Eq:tiled IC}
\end{equation}
where $IC_{t}$ is the tiled ICs; $\lfloor\cdot\rfloor$ is the floor function.
Using \eqref{Eq:tiled IC}, we can obtain the optimal number of ICs that can be calculated in a single cycle for a given PIM array.
According to $IC_{t}$, AR cycles can be calculated by 
\begin{equation}
\scriptstyle
   AR\ cycles\ =\ \big\lceil{\frac{IC}{IC_{t}}}\big\rceil.
\label{Eq:rowcycle}
\end{equation}
When the number of windows in a parallel window increases, the size of tiled OCs decreases because the number of columns of the PIM array is limited, as shown in Fig. \ref{fig:IC1C} (b).
Thus, the tiled OCs size can be obtained by
\begin{equation}
\scriptstyle 
   OC_{t}\ =\ \big\lfloor{\frac{2^Y}{(PW_{w}-K_{w}+1)\times(PW_{h}-K_{h}+1)}}\big\rfloor,
\label{Eq:tiled OC}
\end{equation}
where $2^Y$ is the number of columns of a PIM array.
Similar to AR cycles, according to $OC_{t}$, AC cycles can be calculated by
\begin{equation}
\scriptstyle 
   AC\ cycles\ =\ \big\lceil{\frac{OC}{OC_{t}}}\big\rceil.
\label{Eq:colcycle}
\end{equation}
By plugging \eqref{Eq:rowcycle} and \eqref{Eq:colcycle} into \eqref{Eq:computingcycle2}, we can compute the number of cycles required by the proposed VW-SDK method, formulated as follows:

\begin{equation}
\scriptstyle
    computing\ cycles\ =\ N\ of\ PWs\ \times \ \big\lceil{\frac{IC}{IC_{t}}}\big\rceil \ \times \ \big\lceil{\frac{OC}{OC_{t}}}\big\rceil
\label{Eq:computingcycle3}
\end{equation}

Based on \eqref{Eq:computingcycle3}, we propose an algorithm that finds the optimal channel size and the parallel window shape that minimizes the total computing cycles for a given PIM array and convolutional layer. 
The procedure of the proposed algorithm is described in Algorithm 1 and can be summarized as follows:
\begin{enumerate}
    \item The number of computing cycles $CC_{min}$ is initialized as the required cycles when Im2col mapping is used.

    \item 
    While increasing the parallel window size, calculate the number of the parallel windows and tiled channels ($IC_{t}$ and $OC_{t}$) for the given PIM array.
    
    \item  
    In order to obtain the minimum computing cycles, the number of computing cycles $CC_{vw}$ is calculated by \eqref{Eq:computingcycle3}. 

    \item Then, $CC_{min}$ and $CC_{vw}$ are compared and $CC_{min}$  and $PW_{op}$  are updated.
    If the parallel window size grows to the IFM size, the algorithm terminates and returns the minimum computing cycle, the optimal shape of the parallel window, and the tiled ICs and OCs.
\end{enumerate}

\begin{algorithm}[t]
\caption{VW-SDK Algorithm}
\hspace*{0.2cm}\textbf{Input:}

\hspace*{0.5cm} Network parameters \textbf{I, K, IC, OC}

\hspace*{0.5cm} PIM array size \textbf{$2^{X}, 2^{Y}$}

\hspace*{0.2cm}\textbf{Output: }

\hspace*{0.5cm} Minimum computing cycles $CC_{min}$

\hspace*{0.5cm} Tiled channels $OC_{t}, IC_{t}$

\hspace*{0.5cm} Optimal shape of the parallel window $PW_{op}$ 

\begin{algorithmic}[1]
\label{alg:algorithm1}
\STATE Initialize the computing cycles with im2col

and the parallel window size to the kernel size 

$CC_{min}$ $\leftarrow$ $CC_{im2col}$

$PW_{width}$ $\leftarrow$ $K_{width}$

$PW_{height}$ $\leftarrow$ $K_{height}$

\WHILE{True}
    \STATE {$PW_{width}$ $\leftarrow$ $PW_{width}$ + 1}
    \IF{$PW_{width}$ $>$ width of IFMs}
        \STATE{$PW_{width}$ $\leftarrow$ $K_{width}$ 
        \STATE$PW_{height}$ $\leftarrow$ $PW_{height}$ + 1}
        \IF{$PW_{height}$ $>$ height of IFMs}
            \STATE \textbf{break}
        \ENDIF
    \ENDIF 

    \STATE Calculate $IC_{t}$, $OC_{t}$ by (4) and (6) respectively, 
    
    and computing cycle $CC_{vw}$ by (10).
    \IF{$CC_{min}$ $>$ $CC_{vw}$}
        \STATE Update $PW_{op}$ $\leftarrow$ $PW$ 
        \STATE Update $CC_{min}$ $\leftarrow$ $CC_{vw}$
    \ENDIF 
\ENDWHILE
\RETURN{$IC_{t}$, $OC_{t}$, $CC_{min}$, $PW_{op}$ }
\end{algorithmic}
\end{algorithm}

\section{Experimental Results}
\begin{table}[t]
\centering

\caption{Information of CNNs and Results}
\label{table:algo}
\begin{tabular}{c|c|c|c|c}

\hline\hline
\textbf{$\sharp$} & \textbf{Image} & \textbf{kernel} & \textbf{SDK} & \textbf{VW-SDK} \\
 & \tiny{($I\times I$)} & \tiny{($K\times IC\times OC$)} & \tiny{($PW\times IC\times OC$)} & \tiny{($PW\times IC_{t}\times OC_{t}$)} \\
\hline
\hline
\multicolumn{5}{c}{\textbf{VGG-13}} \\
\hline
1 & 224x224 & 3x3x3x64 & 4x4x3x64 & 10x3x3x64 \\
2 & 224x224 & 3x3x64x64 & 4x4x64x64 & 4x4x64x64 \\
3 & 112x112 & 3x3x64x128 & 4x4x64x128 & 4x4x32x128 \\
4 & 112x112 & 3x3x128x128 & 3x3x128x128 & 4x4x32x128 \\
5 & 56x56 & 3x3x128x256 & 3x3x128x256 & 4x3x42x256\\
6 & 56x56 & 3x3x256x256 & 3x3x256x256 & 4x3x42x256 \\
7 & 28x28 & 3x3x256x512 & 3x3x256x512 & 3x3x256x512 \\
8 & 28x28 & 3x3x512x512 & 3x3x512x512 & 3x3x512x512 \\
9 & 14x14 & 3x3x512x512 & 3x3x512x512 & 3x3x512x512 \\
10 & 14x14 & 3x3x512x512 & 3x3x512x512 & 3x3x512x512 \\
\hline
\multicolumn{3}{c|}{Total cycles} & {114697} & {77102}\\
\hline
\hline
\multicolumn{5}{c}{\textbf{Resnet-18}} \\
\hline

1 & 112x112 & 7x7x3x64 & 8x8x3x64 & 10x8x3x64\\
2 & 56x56 & 3x3x64x64 & 4x4x64x64 & 4x4x32x64 \\
3 & 28x28 & 3x3x128x128 & 3x3x128x128 & 4x4x32x128 \\
4 & 14x14 & 3x3x256x256 & 3x3x256x256 & 4x3x42x256 \\
5 & 7x7 & 3x3x512x512 & 3x3x512x512 & 3x3x512x512 \\
\hline
\multicolumn{3}{c|}{Total cycles} & {7240} & {4294}\\
\hline
\hline
\end{tabular}

\end{table}

\subsection{Settings}
We evaluate our proposed algorithm using VGG13 \cite{vggnet} and Resnet-18 \cite{resnet}, whose layer dimensions are described in Table \ref{table:algo}. 
The proposed algorithm receives the PIM array size and the network parameters of each convolutional layer.
After comparing the computing cycles by Algorithm \ref{alg:algorithm1}, it returns the minimum computing cycles, tiled channels, and optimal shape of the parallel window.
Codes are available at the following address.
\footnote{\url{https://github.com/djwhsdj/VW-SDK}}
 
\subsection{Results}
\begin{figure}[t]
\centering
\includegraphics[trim={0.2cm 0.2cm 0cm 0cm}, clip, width=0.85 \linewidth]{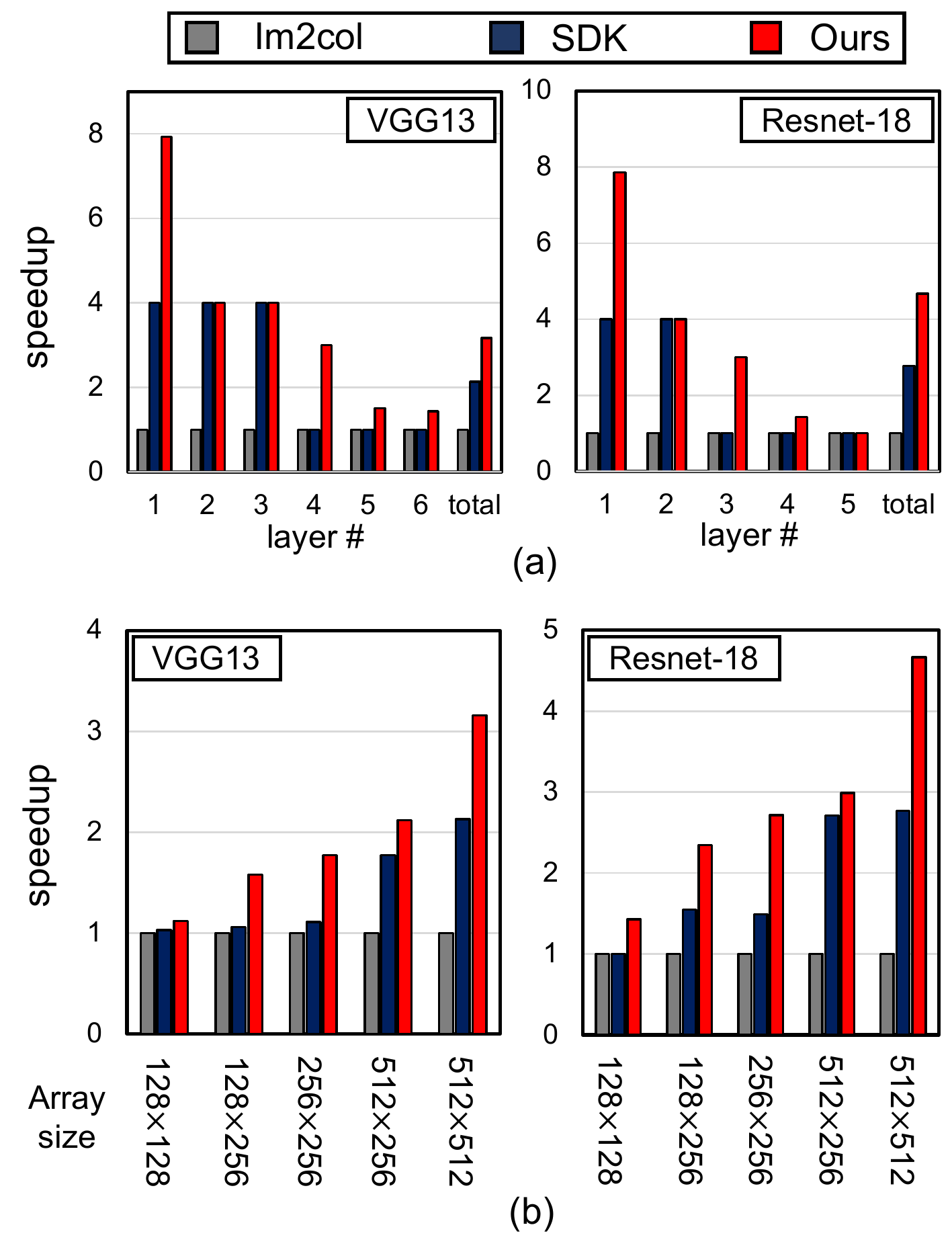}
\caption{Comparison of the speedup normalized to the number of computing cycles of im2col
(a) for each layer of the CNNs using a 512 x 512 PIM array;
(b) for the entire layers of the CNNs according to various PIM array sizes.
}
\label{fig:compare network}
\end{figure}

Table \ref{table:algo} shows the information of the CNNs and the parallel window sizes obtained from the existing SDK-based algorithm and the proposed one, with the 512$\times$512 PIM array.
After the Layer 3 of VGG13 and Resnet-18, Fig. \ref{fig:compare network}(a) shows that the SDK-based algorithm does not reduce the computing cycles from Im2col. This is because it cannot form the parallel window larger than the kernel as the entire channels cannot be unrolled in the given PIM array. 
In contrast, VW-SDK optimizes the parallel window shape that maximizes the utilization of a given PIM array and convolutional layer.
VW-SDK improves the computing speed by 3.16$\times$ and 1.49$\times$ on VGG13, 4.67$\times$ and 1.69$\times$ on Resnet-18 compared to im2col and SDK-based algorithm, respectively.

Fig. \ref{fig:compare network}(b) shows that the speedups of both the SDK-based and the proposed ones increases with larger array size. This is because these algorithms can compute multiple OFMs by reusing the elements of IFMs at a single cycle.
The figure shows that VW-SDK further improves the computing speed by mapping the parallel window with variable channel size and shape.

\begin{figure}[t]
\centering
\includegraphics[trim={0cm 0cm 0cm 0.3cm}, clip, width= 0.95 \linewidth]{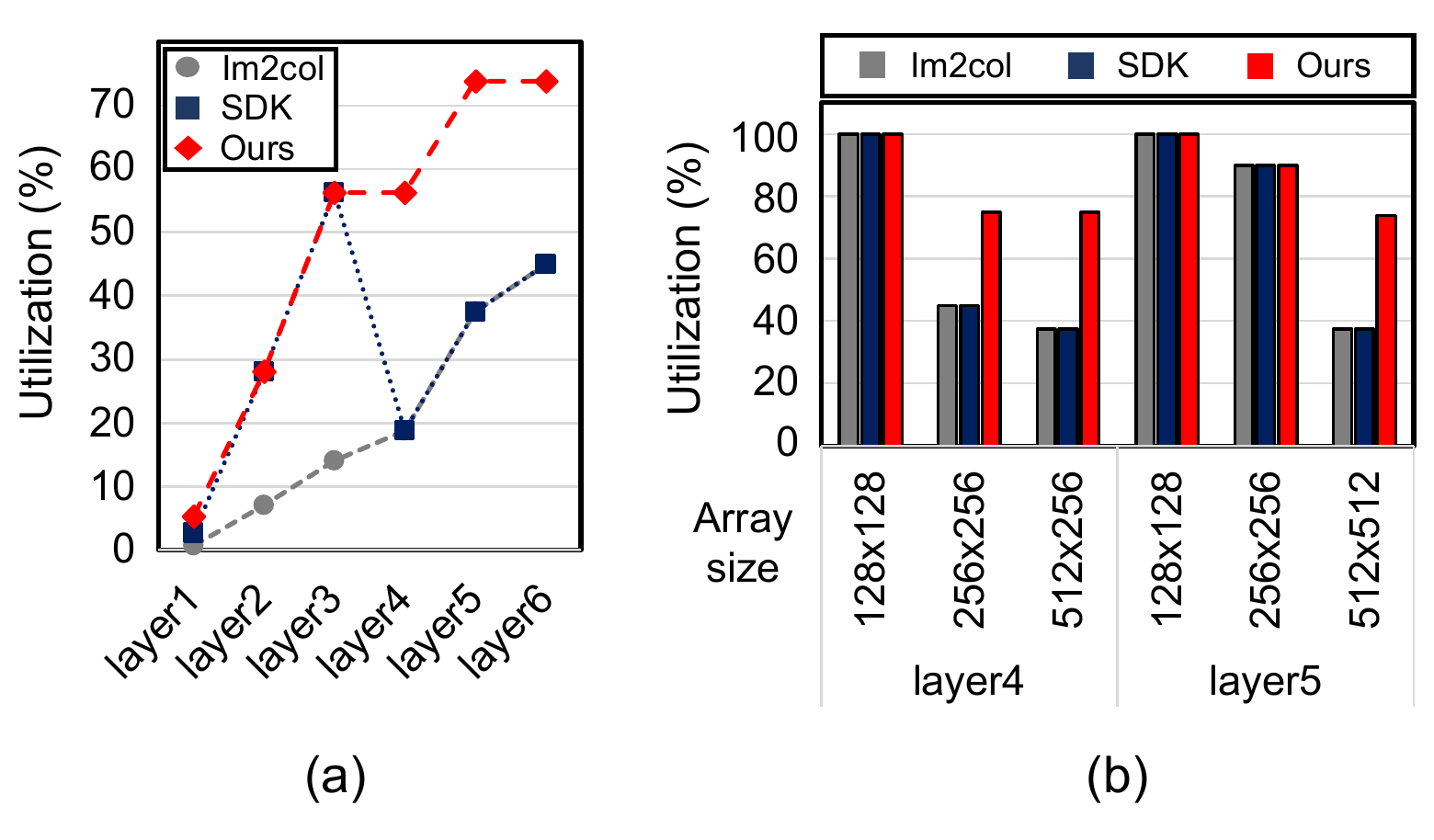}
\caption{
Comparison of the utilization rate in VGG13;
(a) in convolutional layers, where the PIM array size is 512$\times$512.
(b) in the layer4 and layer5 of VGG13 with different PIM array sizes. 
}
\label{fig:utilization}
\end{figure}

We also compare the utilization of the PIM array with different mapping algorithms.
Here, the utilization is defined as the ratio of used memory cells in the PIM array; it is calculated as follows:
\begin{equation}
\scriptstyle 
  Utilization(\%) \ =\ \frac{1}{C}\left(\sum\limits_{n=1}^{C} \frac{U_{n}}{T_{n}}\right) \times 100,
\label{Eq:utilization}
\end{equation}
where $U$ is the number of used memory cells; $T$ is the number of total memory cells in a PIM array; $C$ is cycles including AR and AC cycles. 
Fig. \ref{fig:utilization}(a) shows that the utilizations of the SDK-based algorithm and VW-SDK are equal until Layer 3, as the same shapes of the parallel window are formed.
However, after Layer 3, the SDK-based algorithm fails to form larger parallel windows because the entire channels cannot be mapped to a given array. In contrast, with the tiled channel and rectangular-shaped parallel window, VW-SDK utilizes the given PIM array more efficiently, achieving a utilization up to 73.8\% at Layer 5, where the utilization of other mapping algorithms is only 45\%. 

With a larger PIM array, VW-SDK gains higher utilization than the conventional algorithms, as shown in Fig. \ref{fig:utilization}(b).
This is because regardless of convolutional layers, the proposed algorithm can find a more optimal size of the parallel window by unrolling more tiled channels and shifted and duplicated kernels.
Therefore, we can expect that VW-SDK will be more effectively used as larger PIM arrays are proposed in the future.

\section{Conclusion}
In this paper, we proposed an algorithm called VW-SDK that obtains the optimal shape of a parallel window by forming various windows, leading to the minimum computing cycles for a given convolutional layer and PIM array.
The VW-SDK divides the entire channels of the convolutional layer into several tiles to utilize the SDK mapping method.
Consequently, the number of computing cycles decreases compared to the conventional algorithms regardless of the PIM array size.
We estimated our proposed algorithm with VGG13 and Resnet-18.
VW-SDK improved the computing speed by 4.67$\times$ and 1.69$\times$ on Resnet-18 compared to im2col and the SDK-based algorithm respectively.

\appendices
\section*{Acknowledgment}
This research was supported by the Ministry of Science and ICT (MSIT) of Korea, under the National Research Foundation (NRF) grant (2020M3H2A1076786) and Institute of Information and Communication Technology Planning Evaluation (IITP) grants for the AI Graduate School program (IITP-2019-0-00421), Information Technology Research Center (ITRC) program (IITP-2021-0-02052), and ICT Creative Consilience program (IITP-2020-0-01821).


\end{document}